\documentclass[10pt,twocolumn,letterpaper]{article}

\usepackage{iccv}
\usepackage{times}
\usepackage{epsfig}
\usepackage{graphicx}
\usepackage{amsmath}
\usepackage{amssymb}

\usepackage[ruled]{algorithm2e}
\usepackage{amsmath}
\usepackage{animate}
\usepackage{graphicx}
\usepackage{diagbox}
\usepackage{sidecap}
\usepackage{textcmds}
\usepackage{placeins}
\usepackage{multirow}
\usepackage[normalem]{ulem}
\usepackage{adjustbox}
\usepackage[table,xcdraw]{xcolor}
\usepackage{wrapfig}
\usepackage{caption}
\usepackage{color}
\usepackage{tabularray}
\usepackage{booktabs}

\newcommand{\figref}[1]{Fig.~\ref{#1}}
\newcommand{\tabref}[1]{Tab.~\ref{#1}}

\renewcommand{\eqref}[1]{Eq.~\ref{#1}}

\newcommand\blfootnote[1]{%
  \begingroup
  \renewcommand\thefootnote{}\footnote{#1}%
  \addtocounter{footnote}{-1}%
  \endgroup
}

\usepackage[pagebackref=true,breaklinks=true,letterpaper=true,colorlinks,bookmarks=false]{hyperref}

 \iccvfinalcopy 

\def\iccvPaperID{7430} 

\ificcvfinal\fi

\begin{document}

\title{Task Agnostic Restoration of Natural Video Dynamics}

\author{
Muhammad Kashif Ali
~~~~~~~~
Dongjin Kim ~~~~~~~~
Tae Hyun Kim$^{*}$\\
Department of Computer Science, Hanyang University, Seoul, Korea\\
{\tt\small \{kashifali, dongjinkim, taehyunkim\}@hanyang.ac.kr}
}

\ificcvfinal\thispagestyle{empty}\fi

\twocolumn[{
\maketitle
\begin{center}
    \captionsetup{type=figure}
    \begin{frame}

      \animategraphics[autoplay,loop,width=0.9\linewidth]{20}{video_figures/teaser_3/teaser_mod-}{1}{40}
    
    \end{frame}
    \captionof{figure}{\textbf{Applications of the proposed method.} The proposed algorithm takes only the per-frame processed videos with severe temporal flicker (bottom-left) and produces temporally consistent results (top-right). The proposed method is agnostic to the image processing operations used to process videos, and it does not require the availability of raw/unprocessed videos at test time. The figure presented above contains animated content and is best viewed on a computer screen with Adobe PDF reader.} 
\end{center}
}]

\begin{abstract}
In many video restoration/translation tasks, image processing operations are na\"ively extended to the video domain by processing each frame independently, disregarding the temporal connection of the video frames. This disregard for the temporal connection often leads to severe temporal inconsistencies. State-Of-The-Art (SOTA) techniques that address these inconsistencies rely on the availability of unprocessed videos to implicitly siphon and utilize consistent video dynamics to restore the temporal consistency of frame-wise processed videos which often jeopardizes the translation effect. We propose a general framework for this task that learns to infer and utilize consistent motion dynamics from inconsistent videos to mitigate the temporal flicker while preserving the perceptual quality for both the temporally neighboring and relatively distant frames without requiring the raw videos at test time. The proposed framework produces SOTA results on two benchmark datasets, DAVIS and videvo.net, processed by numerous image processing applications. The code and the trained models are available at \url{https://github.com/MKashifAli/TARONVD}.\blfootnote{\hspace{-2em}$^*$: Corresponding author.}
\end{abstract}


\section{Introduction}
Video sharing social media platforms like Snapchat and TikTok have introduced the common populace to a plethora of computer vision applications such as Style Transfer~\cite{gatys2015neuralstyletransfer}, Colorization~\cite{zhang2016imagecolorization}, Denoising~\cite{zhang2017dncnn}, and Dehazing~\cite{yan2016imageadjustment}. With this wide-scale integration of these classical computer vision applications in such platforms, various image processing operations are na\"ively extended to videos due to scarcity of annotated video datasets and their computational complexity of video processing methodologies. This na\"ive extension of static image processing methodologies to videos disregards the temporal connection of the video progression and introduces severe temporal flickering in the videos. This temporal flicker can appear for various reasons; for instance, these image processing methods can produce drastically different results for temporally neighboring frames due to slight changes in their global or local content statistics, or it could also happen due to the multi-modality of the application as there could exist a number of valid solutions for images with similar content as highlighted in~\cite{bonneel2015blind, zhang2019deepexemplarcolorization}. Therefore, the extension of these image-to-image translation tasks to the video domain is an active area of research in computer vision, and methods that can help extend various image processing applications with little to no knowledge of the operations used to create videos are quite useful.

Currently, there are several task-dependent temporal consistency correction approaches available such as~\cite{chen2017videostyletransfer, liu2021featurepropagation, thimonier2021loreal}, but due to the complex nature of these applications, only a handful of approaches have been proposed to tackle the problem of blind temporal consistency correction. Bonneel et al.~\cite{bonneel2015blind} initiated the investigation of blind temporal consistency correction with gradient-domain minimization of per-frame processed video with the unprocessed video to minimize the warping error between the frames. Lai et al.~\cite{lai2018learning} extended the formulation of~\cite{bonneel2015blind} with the help of recurrent  Convolutional Neural Networks (CNN) and introduced a perceptual penalty in their formulation to restrict the deviation of perceptual content of the restored video from the frame-wise processed video. Deep Video Prior (DVP)~\cite{lei2020blind} extended Deep Image Prior (DIP)~\cite{ulyanov2018deepimageprior} to the temporal dimension and proposed to formulate enforcing temporal consistency by training a CNN to generate processed video from the unprocessed video without utilizing optical flow. All of these approaches rely on the availability of unprocessed videos for implicit extraction of consistent motion dynamics to use as a restoration guide. 
Please note that all the previously proposed approaches for this task define it with the help of unprocessed videos (both task-specific and task-agnostic). Having access to unprocessed videos at test time helps the models in siphoning and transferring consistent motion dynamics to the inconsistent videos to decrease temporal flicker. Despite its efficacy, this implicit definition limits the applicability of the previously proposed approaches to only the videos for which their unprocessed counterparts are available and also introduce an inherent bias towards the unprocessed videos which can compromise the translation effect and quality of the processed videos (as presented in Fig.~\ref{bonvsours} $\sim$ \ref{laivsour}). 

In order to overcome these limitation, we aim to learn and infer consistent motion representation solely from temporally inconsistent videos. Doing so does not only alleviate the need for raw videos at test time, but also mitigates the inherent bias towards the raw videos that is commonly present in the currently available approaches for this task.
For the task of learning these consistent motion representation, we fine-tune conventional optical flow estimations networks like~\cite{ilg2017flownet, sun2018pwc, teed2020raft} along with the restorative part of the network in an end-to-end manner. The proposed model pipeline is illustrated in \figref{main_figure}. The proposed model achieves state-of-the-art qualitative and quantitative results. The proposed formulation can also handle the resolution mismatch problem in processed and raw videos and makes it possible to extend the Single Image Super-Resolution~\cite{ledig2017sisr} methods to Video Super-Resolution methods without any modification. The detailed description of our formulation is presented in \ref{methodology}. We summarize our contributions as follows:\\
$\bullet$ Identify and propose tailored solutions for various challenges of na\"ive extensions of image processing applications to videos.\\
$\bullet$ Learn to infer consistent motion representations from temporally inconsistent videos.\\
$\bullet$ Utilize the learned consistent motion representations to propose a general framework for task agnostic temporal consistency correction that produce SOTA results.


\section{Related Work}
The literature is divided into two main streams to generate visually appealing videos through image transformation models. The first stream tackles the task of mitigating temporal inconsistencies with the reformulation of the task at hand. Whereas, The second stream focuses on developing post-processing models that refine the frame-wise processed videos by penalizing the temporal deviation of the processed video from its unprocessed counterpart. The second stream is further divided into two sub-streams: task-specific and task-agnostic. The following passages summarize the details of both of the above-mentioned streams, respectively.\\
\textbf{Reformulation stream (video-to-video translation):}
	These approaches are generally termed video-to-video translation applications. Generally, these approaches either consider multiple frames as input and produce multiple output frames or generate a single frame from multiple input frames in the form of frame recurrence such as~\cite{chu2018tecogan, liu2021featurepropagation}. There are also cases like video style transfer where content information is propagated to the next time step with the help of optical flow to initiate the optimization of the next frame. This frame-recurrent methodology has also been proven effective in applications like video super resolution~\cite{FRVSR}. Designing these approaches for each task and training them from scratch is a taxing task and data scarcity can make these formulations unfeasible. Generally, such models often do not adapt well to different tasks; therefore, methods that can restore the temporal consistency of multiple frame-wise processed videos are being actively investigated.\\ 
\textbf{Post-processing methods:}
	Bonneel et al.~\cite{bonneel2015blind} initiated the investigation of task agnostic temporal consistency correction using a gradient-based optimization strategy in which temporally consistent (unprocessed) videos were used as restoration guides for correcting the temporal inconsistency of frame-wise processed videos. Their formulation motivated various task-dependent and task-agnostic, and temporal consistency correction approaches with slight variations of their original objective formulation. Despite the simplicity of their formulation, there only exist a handful of approaches that address task-agnostic temporal consistency correction due to the difficult nature of the task. Yao et al.~\cite{yao2017occlusion} proposed a key-frame strategy that accounted for the motion of different objects in those key-frames to handle occlusion as well as temporal consistency. Lai et al.\cite{lai2018learning} proposed the first deep learning approach for this task by employing ConvLSTM~\cite{shi2015convlstm} and a perceptual loss~\cite{johnson2016perceptualloss}. Lei et al.~\cite{lei2020blind} proposed an extension of Deep Image Prior~\cite{ulyanov2018deepimageprior} and demonstrated its capability to mitigate the temporal flicker of per-frame processed videos.

Apart from these general frameworks, various task-specific approaches have also been proposed, such as~\cite{chu2018tecogan, liu2021featurepropagation, thimonier2021loreal}. Defining these task-specific approaches is relatively straightforward. These approaches define a temporal extension around a backbone method and penalize the deviation of generated content with the help of motion information from the unprocessed videos.

In this work, we propose a general formulation for this task that alleviates the requirement of unprocessed videos at test time by learning to infer a consistent motion representation from the temporally inconsistent videos. This inferred motion representation is then processed with the content of temporally inconsistent videos in a two branched network that restores the natural dynamics of per-frame processed videos. Our proposed network also contains a recurrent bottleneck module that helps consistent propagation of content throughout the video sequence and produces SOTA qualitative and quantitative results by only taking into account the temporally inconsistent videos at test time.

In order to aid the readers, we provide the following table to highlight the key differences between the currently available approaches for this task.

\begin{table}[h]
\adjustbox{max width=0.48\textwidth}{
  \centering
    \begin{tabular}{|l|cccc|}
    \hline
    \multicolumn{1}{|c|}{\textbf{Key features}} & \multicolumn{1}{c}{\textbf{Bonneel et al.~\cite{bonneel2015blind}}} & \multicolumn{1}{c}{\textbf{Lai et al.~\cite{lai2018learning}}} & \multicolumn{1}{c}{\textbf{DVP~\cite{lei2020blind}}} & \textbf{Ours} \\
    \hline
    Does not require raw videos & \textcolor{red}{$\times$} & \textcolor{red}{$\times$} & \textcolor{red}{$\times$} & \textcolor{green}{\textbf{$\checkmark$}} \\
    Explicit video dynamics design & \textcolor{green}{\textbf{$\checkmark$}} & \textcolor{red}{$\times$} & \textcolor{red}{$\times$} & \textcolor{green}{\textbf{$\checkmark$}} \\
    Faithful perceptual restoration & \textcolor{red}{$\times$} & \textcolor{red}{$\times$} & \textcolor{red}{$\times$} & \textcolor{green}{\textbf{$\checkmark$}} \\
    Sequential processing & \textcolor{green}{\textbf{$\checkmark$}} & \textcolor{green}{\textbf{$\checkmark$}} & \textcolor{red}{$\times$} & \textcolor{green}{\textbf{$\checkmark$}} \\
    Long term consistency & \textcolor{red}{$\times$} & \textcolor{red}{$\times$} & \textcolor{green}{\textbf{$\checkmark$}} & \textcolor{green}{\textbf{$\checkmark$}} \\
    Sharpness in processed videos & \textcolor{red}{$\times$} & \textcolor{green}{\textbf{$\checkmark$}} & \textcolor{red}{$\times$} & \textcolor{green}{\textbf{$\checkmark$}} \\
    High-frequency flicker removal & \textcolor{green}{\textbf{$\checkmark$}} & \textcolor{red}{$\times$} & \textcolor{green}{\textbf{$\checkmark$}} & \textcolor{green}{\textbf{$\checkmark$}} \\
    Test time training & \textcolor{red}{$\times$} & \textcolor{red}{$\times$} & \textcolor{green}{\textbf{$\checkmark$}} & \textcolor{red}{$\times$} \\
    Runtime (per frame avg.) & 2.4635* & 0.2146 & 3.6365 & 0.2236 \\
    \hline
    \end{tabular}}
    \caption{\textbf{Key differences in available approaches.} "*" in front of Bonnel et al.~\cite{bonneel2015blind}'s runtime highlights environment difference and is only presented for reference only. Please note that the runtime is averaged over the same video sequence for all the methods.}
  \label{explicit_difference}%
  \vspace{-10pt}
\end{table}%
\section{Proposed Method} \label{methodology}
\begin{figure}[ht]
\begin{center}
\includegraphics[width=1.0\linewidth]{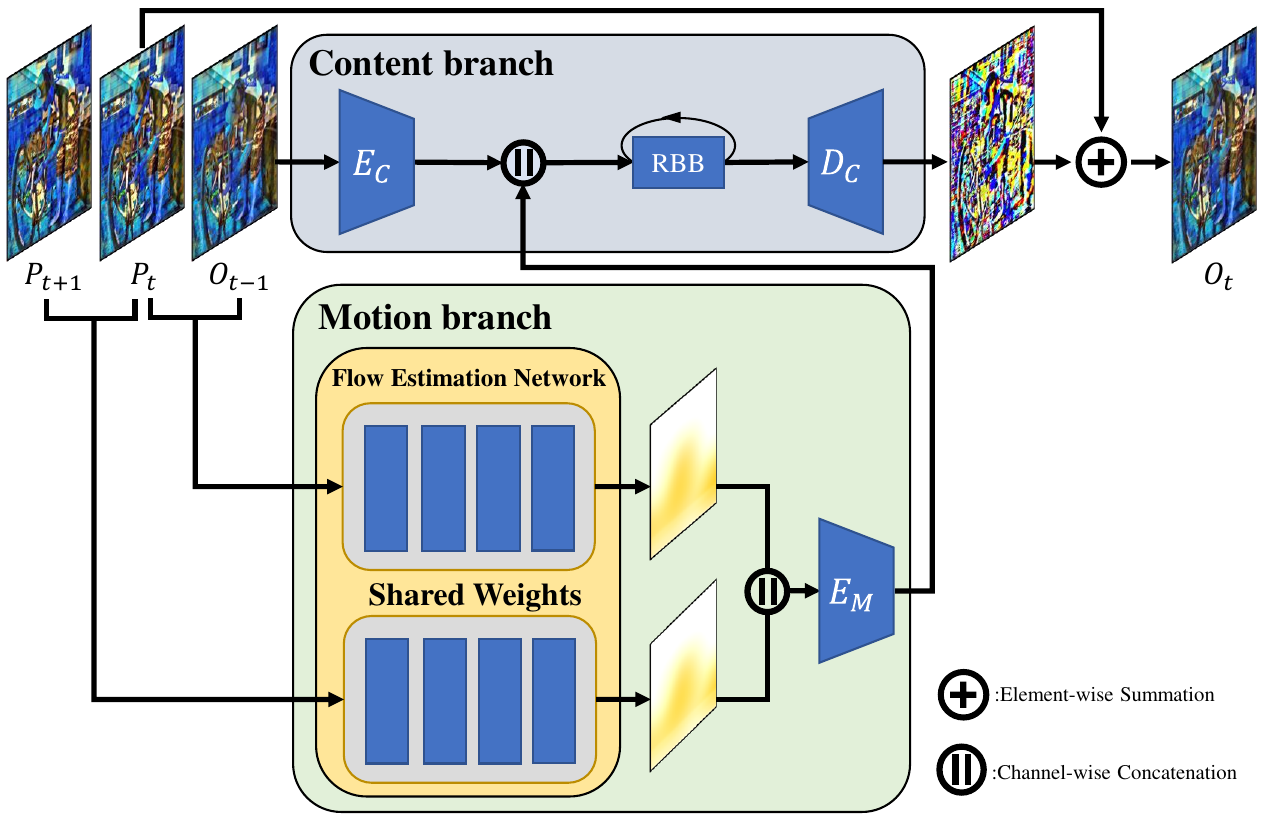}
\end{center}
\vspace{-15pt}
\caption{\textbf{Model Architecture.} An illustration of the proposed model architecture. The proposed model consists of a two-branch architecture that separately processes motion and content. The combined motion and content features concatenated along the channel dimension and are then passed through a Recurrent Bottleneck Block (RBB) and a decoder to generate the restored frame.} 
\vspace{-15pt}
\label{main_figure}
\end{figure}

In this section, we describe the details of the proposed method. Consider a raw (temporally consistent) video \{$I_{0}$, $I_{1}$, $I_{2}$, ..., $I_{n}$\} with $n$ frames and its frame-wise processed (temporally inconsistent) video \{$P_{0}$, $P_{1}$, $P_{2}$, ..., $P_{n}$\} acquired by an image processing function $h$ as \{$h(I_{0})$, $h(I_{1})$, $h(I_{2})$, ... , $h(I_{n})$\}. The goal of this task is to restore the temporal consistency of the temporally inconsistent video and to produce a temporally consistent version \{$O_{0}$, $O_{1}$, $O_{2}$, ..., $O_{n}$\}. For the sake of simplicity, these notations are chosen to be in line with the notations used in past literature~\cite{bonneel2015blind, lai2018learning, lei2020blind}.

The proposed formulation draws inspiration from the disentanglement of a video into its base components: content and motion representations as presented in~\cite{lin2017disentangling}. In their work~\cite{lin2017disentangling}, they investigated the controlled generation of short length videos by mixing motion and content components from various videos. In this work, we explore a similar concept of controlled disentanglement and re-entanglement strategy to restore natural video dynamics of the frame-wise processed videos. More specifically, we argue that the previously proposed approaches for this tasks implicitly employ this disentanglement and re-entanglement in their formulation with the help of unprocessed video frames. Conventionally, optical flow is used as an inter-frame motion representations. The estimated optical flow can also be used to approximate the content of future frames with the help of past frames provided that there does not exist occlusion between the frames (as presented in Eq.~\ref{warping_eq}). 
\begin{equation} \label{warping_eq}
\small
\hat{I}_{t+1} \approx w\left(I_{t}, o f_{t+1 \Rightarrow t}\right).    
\end{equation}

\begin{figure}[ht]
\begin{center}
\includegraphics[width=1.0\linewidth]{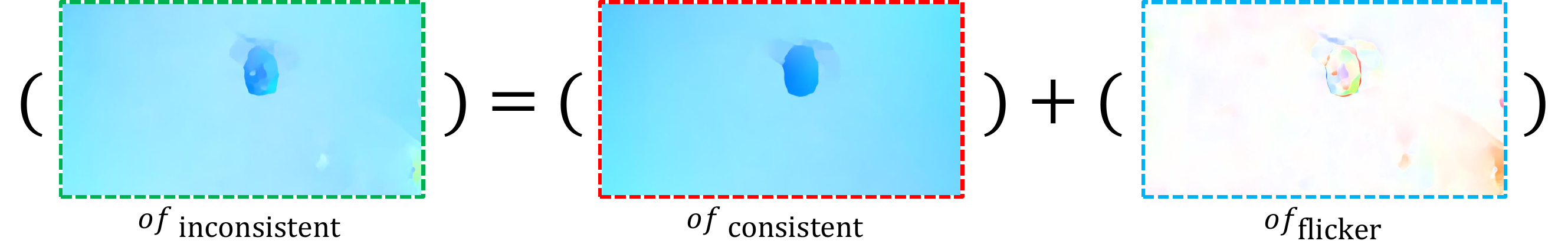}
\end{center}
\vspace{-10pt}
\caption{\textbf{Temporal flicker in optical flow.} An illustration of the decomposition of optical flow generated from temporally inconsistent frames to consistent optical flow and inconsistent flicker.} 
\vspace{-11pt}
\label{flowplusnoise}
\end{figure}
This interdependence of optical flow and content can lead to inconsistent motion representation as optical flow estimation networks are not trained explicitly to distinguish temporally consistent content from temporal flicker. Due to this limitation, the evaluated optical flow estimated from temporally inconsistent frames is often compromised with temporal flicker (as presented in Fig.~\ref{flowplusnoise}). This compromise of the motion information makes it challenging to separate the consistent component of optical flow from the additive flicker. Therefore, the currently available techniques for this task often rely on siphoning implicit motion representations from the raw videos and often suffer in cases where the content of both processed and raw videos deviate significantly. Another problem with employing these implicit representations is the bias introduced due to the lack of control in mixing these implicit representations to restore videos which often leads to the mitigation of translation effect. Therefore, if a mechanism can be developed that produces similar motion representations from both unprocessed and frame-wise processed videos, better and unbiased restorative effects than those produced by the previously proposed approaches can be achieved without relying on the availability of unprocessed videos.

Please note that another observation that can be made from \figref{flowplusnoise} (right most figure) is that a significant amount of flicker in frame-wise processed videos, generally occurs near the motion boundaries. This observation helps us in defining objective functions capable of localizing and targeting temporal deviations as presented in Sec.~\ref{fg_function}. 

To overcome these problems and to learn and utilize consistent motion representation, we integrate an optical flow estimation network into the restoration network as illustrated in \figref{main_figure}. A similar integration of a small-scale optical flow estimation network (SpyNet~\cite{ranjan2017spynet}) for task-oriented flow estimation is presented in~\cite{xue2019toflow} for the task video denoising for its extremely small size. We utilize a relatively larger network (PWC-Net~\cite{sun2018pwc}) in our method and provide an ablation study on different sized optical flow models in the accompanying supplementary text. We further extend their idea to a two pass mechanism with an added encoder to encourage generation of consistent dynamics. The optical flow network in the proposed methodology is finetuned in an end-to-end manner without any special supervision. 

The proposed model takes in three frames $O_{t-1}$ ($P_{t-1}$ for the first time step), $P_{t}$ and $P_{t+1}$ as input. These three frames are propagated through both the content and motion branches. The proposed model consists of a UNet~\cite{ronneberger2015unet} like structure with multiple encoder streams and a single decoder stream. The decoder contains skip connections from the content stream to encourage better reconstruction. The bottleneck part of the model contains a recurrent bottleneck block to transmit and manipulate the information of the generated frames to temporally distant frames.
Please note that the multi-branch encoder(s)-bottleneck-decoder architecture used for the proposed model is employed due to its proven efficacy in video generation, translation and restoration tasks as described in~\cite{chen2017videostyletransfer, Gao2020FastVM, Kim2019DeepBV, Kim2019RecurrentTA, lai2018learning, liu2021featurepropagation, sun2020twostreamvan, thimonier2021loreal}. The main difference in the proposed architecture comes from the design of our motion branch which allows us to present a standalone solution for this task. 
The motion branch comprises of two passes through a PWCNet~\cite{sun2018pwc} with shared weights followed by a conventional encoder-like architecture. An illustration of the proposed model is presented in \figref{main_figure}. This two-pass strategy in the motion branch along with the attached encoder allows the proposed model in regressing consistent motion dynamics from temporally inconsistent frames. Further architectural details of the proposed network are presented in the accompanying supplemental. 
\begin{figure}[t]
\begin{center}
\includegraphics[width=0.8\linewidth]{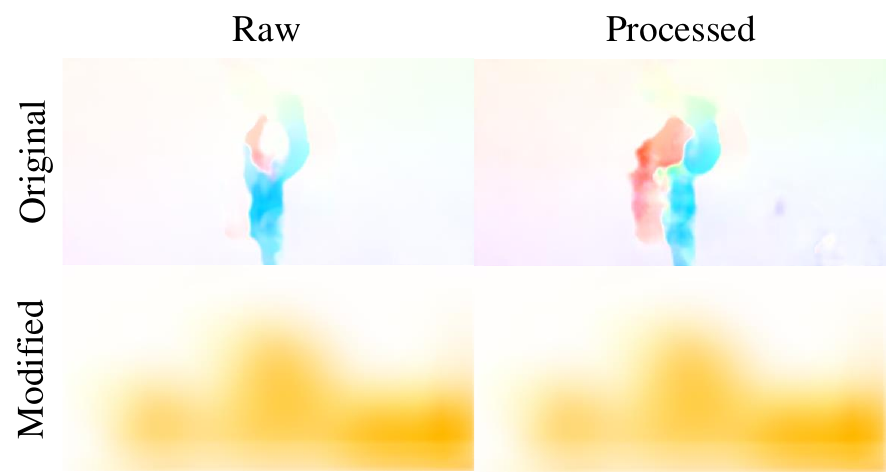}
\end{center}
\vspace{-10pt}
\caption{\textbf{Conventional and Modified flow like representation.} Comparison of original and Modified flow evaluated from temporally consistent (left) and inconsistent (right) frames. The modified flow evaluated from both the temporally consistent and inconsistent frames is quite similar.}
\vspace{-13pt}
\label{modifiedflow_figure}
\end{figure}
Figure~\ref{modifiedflow_figure} presents conventional optical flow along with the learned motion representation generated by the integrated finetuned PWC-Net~\cite{sun2018pwc} from temporally consistent (left) and inconsistent frames (right). Please note that the "modified" flow can differ starkly from the conventional flow and can only be interpreted effectively by the network it is trained with as described in~\cite{xue2019toflow}. Contrasting to the conventional optical flow modified flow estimated from both consistent and inconsistent videos is quite similar, which suggests that the network can effectively utilize the inferred consistent motion representation to restore natural temporal dynamics in the per-frame videos by only taking into account the temporally inconsistent videos at test time. Another benefit of this disentangled representation is that it allows the users to further improve the level of restored temporal consistency with the help of an iterative arrangement; please refer to the supplemental for the details of the iterative arrangement and the results generated with such arrangement. The objective functions used for the end-to-end training of the proposed network are described in detail in the next section. 

\section{Objective Functions}
This section provides the details of the learning objectives used in the training phase of the proposed network. These objectives can be classified into two categories, local neighborhood and long-term objectives. Both of these optimization categories are presented below.

\subsection{Local Neighborhood Losses} \label{fg_function}
In this work, we aim to find a general solution for correcting the temporal consistency of frame-wise processed videos. Due to the diversity of the applications addressed, a simplistic spatial content matching reconstruction loss cannot justify the generation of temporally consistent frames, as the content can vary significantly in applications like style transfer~\cite{gatys2015neuralstyletransfer} and image adjustment~\cite{yan2016imageadjustment}. This discrepancy in the image space of processed and raw videos hinders a straightforward definition of a feasible loss function for this task. Due to the challenging nature of this task, we propose a flow gradient loss that provides a supervision signal for the reconstruction of motion boundaries (where most of the inconsistencies/flicker appear; as highlighted in Fig.~\ref{flowplusnoise}) with the help of optical flow acquired through the unprocessed videos. We further extend this loss to only compare the gradients of the optical flow acquired through the synthesized and raw frames. Doing so reduces the redundancy present in the high dimensional optical flow which contains a redundant global component for this task which is only useful in applications like video stabilization~\cite{ali2020deep} as described in~\cite{yang2021glm}. The proposed flow gradient loss encapsulates the local spatio-temporal information necessary to mitigate temporal flicker near the motion boundaries by removing the unnecessary high-dimensional components from the optical flow. Similar information bottle-necking techniques have been discussed thoroughly in vision based understanding and manipulation literature~\cite{dwibedi2020counting, ma2020structure}. The proposed spatio-temporal loss is defined as follows:
\vspace{-5pt}
\begin{equation}\label{fg_loss}
\small
\mathcal{L}_{fg}=\sum_{t=2}^{T} \left\|\nabla\left(of\left(O_{t}, O_{t - 1}\right)\right), \nabla\left(of\left(I_{t}, I_{t - 1}\right)\right)\right\|_{1}.
\end{equation}

Here, $\mathcal{L}_{fg}$ and $of$ represents flow gradient loss and the optical flow estimation network (FlowNet2.0~\cite{ilg2017flownet}) respectively. Please note that any optical flow estimation network (SpyNet~\cite{ranjan2017spynet}, PWC-Net~\cite{sun2018pwc}, RAFT~\cite{teed2020raft}, etc.) can be used in both the motion branch and the definition of the loss functions, we choose FlowNet2.0~\cite{ilg2017flownet} for its relatively lower mean end-point-error, as the estimated optical flow can vary significantly for small scale networks (comparison provided in the supplementary text). In equation~\ref{fg_loss}, the $\nabla$ operator denotes the spatial gradient of the estimated optical flow. 
A similar loss in optical flow domain has been proposed in~\cite{li2022inpainting} where the raw optical flow of the generated video was compared with that of the ground truth video. Although effective for the task of video inpainting, the raw optical flow matching approach fails in applications like style transfer in which the entire content of the generated video and the raw video differ greatly, i.e., contains excessive spatio-temporal variations which contribute to optical flow estimation. In our experiments, the models trained with raw optical flow performed poorly, whereas the model trained with flow gradients converged in a relatively shorter time and produced sharper frames.

Please note that this optical flow-based loss by itself is not sufficient for faithfully correcting the temporal consistency, as there could exist multiple solutions for optical flow equation, e.g., it does not take into account the appearance of the synthesized and raw frames. In order to address this issue and to guide the network for the generation of perceptually credible frames, a non-local optical flow based loss in the image space is employed as presented below:
\vspace{-3pt}
\begin{equation} \label{st_loss}
\vspace{-5pt}
\small
\mathcal{L}_{recon}=\sum_{t=2}^{T} C_{t}\left\|O_{t}-\hat{O_{t}}\right\|_{1},
\end{equation}

This warping based reconstruction loss function has been adopted from the~\cite{chen2017videostyletransfer}, as many variants of this objective function with slight modifications have been proven effective for enforcing temporal smoothness in video translation tasks in~\cite{bonneel2015blind, chu2018tecogan, lai2018learning, liu2021featurepropagation, thimonier2021loreal}. Here $T$ represents the total number of frames in a sequence and $\hat{O_{t}}$ represents the warped frame $O_{t-1}$ (as described in Eq. \ref{warping_eq}). The working of this loss function can be summarized as the propagation of content from preceding frames to the subsequent frames in image domain to ensure that the generated frames contain similar content to the previous frames and the inter-frame motion dynamics of the synthesized frames are similar to that of the temporally consistent counterpart. 
This non-local optical flow based loss function also takes into account the occlusion problem of the Eq.~\ref{warping_eq} by masking the occluded and de-occluded content as defined below:

\begin{equation}
\small
C_{t} = \exp \left(-\alpha\left\|I_{t}-\hat{I_{t}}\right\|_{2}^{2}\right).
\end{equation}

The value for $\alpha = 50$ is chosen according to the previous works where a similar strategy is used for evaluating occlusion~\cite{thimonier2021loreal, ruder2016artistic}.

An additional conventional short-term perceptual similarity loss~\cite{johnson2016perceptualloss} was also introduced in the training phase to minimize the deviation of the synthesized frames from the original processed frames. This loss is defined as follows:

\begin{equation}
\small
\mathcal{L}_{p}= \sum_{t}\sum_{l}\left\|\phi_{l}\left(O_{t}\right)-\phi_{l}\left(P_{t}\right)\right\|_{1},
\end{equation}
Here $\phi_{l}(.)$ represents layers of a VGG-16 network till the layer $relu\_4\_3$ (trained on the ImageNet dataset~\cite{deng2009imagenet}).

\subsection{Temporal Constancy Loss}
The loss defined in Eq.~\ref{st_loss} adequately defines the content propagation from the previous frame to the current frame but lacks the ability to enforce it in temporally distant instances. For instance, consider the task of colorization where a car appears in a portion of the frames and is assigned some random color, and it disappears and re-appears in a later instance. The frame-wise colorization method can assign it a different color in each sequence, and the loss function defined in Eq.~\ref{st_loss} will be sufficient to enforce the color constancy of the car in each interval where the car is visible. This, however, does not ensure the constancy of the color assigned to the car in both intervals, i.e., the car could be assigned a solid blue color in the first interval and a solid red color in the next interval which is quite prominent in methods like~\cite{lai2018learning, thimonier2021loreal}. To address this problem of temporally distant yet similar instances, we introduce a recurrent module (ConvLSTM~\cite{shi2015convlstm}) in our proposed model and add explicit constraints that force the model to generate temporally consistent frames. 
The previously proposed approaches for this task simply try to minimize the deviation of current frames from the first frame of the sequence, which is inadequate for videos containing largely varying frames. This variation of the content can render this loss ineffective, and the inconsistencies can only be weakly penalized with the help of motion dynamics siphoned from the raw videos. Therefore, we introduce a bi-directional penalty that enforces the temporal smoothness in both backward (through the local neighborhood losses) and forward (through the proposed long-term loss) direction. This bi-directional optimization strategy is inspired from the ping-pong loss dynamic as presented in~\cite{chu2018tecogan}. The proposed long-term loss is termed as constancy loss and is presented below:

\vspace{-15pt}
\begin{equation} \label{lt_loss}
\vspace{-5pt}
\small
\mathcal{L}_{constancy}=\sum_{p=1}^{T-2}\sum_{t=p+2}^{T}  C_{t \Rightarrow p}\left\|O_{t}-w({O}_{p}, of(I_{t-1},  I_{p}))\right\|_{1}.
\end{equation}

Here subscript $p$ highlights that the preceding distant images are compared with all the subsequent frames. Unlike the simplistic long-term temporal constraints used in~\cite{lai2018learning, thimonier2021loreal} which only penalized the deviation of content from the first frame, this loss term takes into account the inter-frame flicker of all the frames in a sequence and provides a strong supervision signal capable of penalizing temporal discrepancies throughout the sequence. Due to this stronger temporal penalty accompanied with the proposed flow gradient loss, the proposed network learns to produce smoother videos despite having access to only the frame-wise processed videos at test time. The final loss for each training sequence is given by,

\vspace{-15pt}
\begin{equation}
    \mathcal{L}_{total} = \lambda_{1}\mathcal{L}_{fg} + \lambda_{2}\mathcal{L}_{recon} + 
    \lambda_{3}\mathcal{L}_{p} + 
    \lambda_{4}\mathcal{L}_{constancy}. 
\end{equation}

The $\lambda$(s) in the equation above define the contribution of each loss in the optimization phase. The details of these hyperparameters are provided in the supplementary text. 

\section{Experiments and Results} \label{exps}
\begin{figure}[ht]
\begin{center}
\vspace{-0.05\linewidth}
\includegraphics[width=0.9\linewidth]{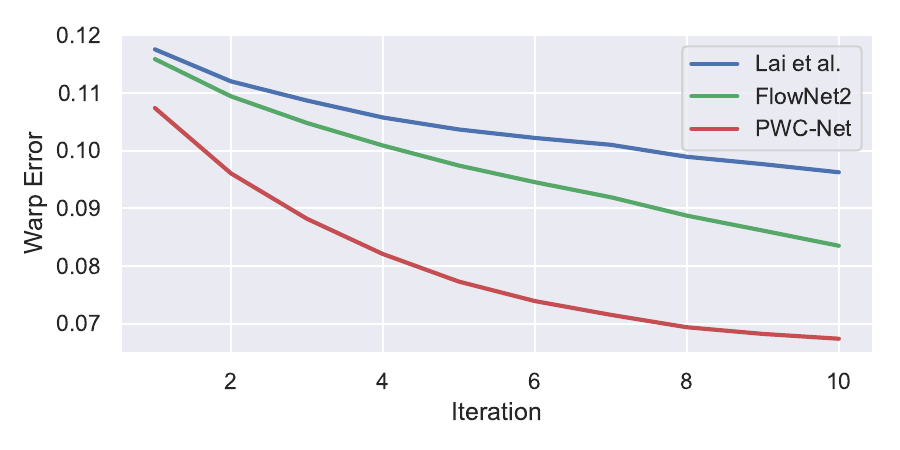}
\end{center}
\vspace{-0.08\linewidth}
\caption{\textbf{Temporal warp error vs. iterations.} Temporal warp error consistently decreases with a higher number of restoration iterations.} 
\vspace{-12pt}
\label{iteration_figure}
\end{figure}

During our experimentation phase, we tested various optical flow estimation networks such as~\cite{ilg2017flownet, ranjan2017spynet, sun2018pwc}. Flownet2.0~\cite{ilg2017flownet} has approximately 162M trainable parameters, and it produced relatively better results in lesser training iterations as compared to the other two variants. Having a vast number of parameters compromises both the training and testing time. Therefore, we opted for a medium-sized optical flow estimation network (PWC-Net~\cite{sun2018pwc}) and a higher number of optimization iterations. The results produced by both variants are discussed in the supplementary text. We also experimented with an iterative arrangement of the trained models where restored videos are again subjected to temporal consistency correction models and evaluated that the proposed model architecture consistently reduces warp error with every iteration (as presented in \figref{iteration_figure}). 
Generally, the frames subjected to a high number of consistency correction iterations lose fine details. The generated results by various restoration iterations are provided in the supplemental. Please note that the comparative results provided in this paper are generated through a single restoration iteration.


The dataset for training the proposed model contains videos from the DAVIS dataset for video segmentation~\cite{perazzi2016davis} and videos gathered from Videovo.net by~\cite{lai2018learning}. The training dataset contains videos processed through a diverse range of applications such as, Artistic Style Transfer~\cite{ gatys2015neuralstyletransfer, johnson2016perceptualloss}, Colorization~\cite{waechter2014let, zhang2016colorful}, Image Enhancement~\cite{gharbi2017deep}, Intrinsic Image Decomposition~\cite{bell2014intrinsic}, and Image-to-Image Translation tasks~\cite{ li2017universal, zhang2016colorful, zhu2017unpaired}. The qualitative and quantitative results are described below.

\subsection{Quantitative Results}
\begin{table}[h]
\adjustbox{max width=0.48\textwidth}{
  \centering
    \begin{tabular}{|l|ccc|c|ccc|c|}
    \hline
    \multicolumn{1}{|c|}{\multirow{2}[2]{*}{Task}} & \multicolumn{4}{c|}{DAVIS}    & \multicolumn{4}{c|}{VIDEVO} \\
    \multicolumn{1}{|c|}{} & Bonneel\cite{bonneel2015blind} & Lai\cite{lai2018learning}   & \multicolumn{1}{c}{Ours} & DVP\cite{lei2020blind}* & Bonneel\cite{bonneel2015blind} & Lai\cite{lai2018learning}   & \multicolumn{1}{c}{Ours} & DVP\cite{lei2020blind}* \\
    \hline
    WCT~\cite{li2017universal}/antimono                                        & \textcolor{blue}{0.0029} & 0.0031 & \textcolor{red}{0.0026} & 0.0022 & \textcolor{blue}{0.0015} & 0.0021 & \textcolor{red}{0.0015} & 0.0017 \\
    WCT~\cite{li2017universal}/asheville                                       & \textcolor{blue}{0.0047} & 0.0059 & \textcolor{red}{0.0047} & --      & \textcolor{red}{0.0032} & 0.0043 & \textcolor{blue}{0.0039} &  --\\
    WCT~\cite{li2017universal}/candy                                           & \textcolor{red}{0.0034} & 0.0047 & \textcolor{blue}{0.0035} & --      & \textcolor{red}{0.0020} & 0.0032 & \textcolor{blue}{0.0025} &  --\\
    WCT~\cite{li2017universal}/feathers                                        & 0.0040 & \textcolor{blue}{0.0040} & \textcolor{red}{0.0027} & --      & \textcolor{blue}{0.0027} & 0.0030 & \textcolor{red}{0.0021} &  --\\
    WCT~\cite{li2017universal}/sketch                                          & 0.0036 & \textcolor{blue}{0.0029} & \textcolor{red}{0.0021} & --      & 0.0025 & \textcolor{blue}{0.0022} & \textcolor{red}{0.0017} &  --\\
    WCT~\cite{li2017universal}/wave                                            & \textcolor{blue}{0.0033} & 0.0035 & \textcolor{red}{0.0027} & 0.0026 & \textcolor{blue}{0.0023} & 0.0026 & \textcolor{red}{0.0021} & 0.0019 \\
    Fast-neural-style~\cite{johnson2016perceptualloss}/princess                & \textcolor{blue}{0.0043} & 0.0063 & \textcolor{red}{0.0042} & 0.0038 & \textcolor{red}{0.0035} & 0.0053 & \textcolor{blue}{0.0042} & 0.0046 \\
    Fast-neural-style~\cite{johnson2016perceptualloss}/udnie                   & \textcolor{red}{0.0021} & 0.0023 & \textcolor{blue}{0.0022} & 0.0022 & \textcolor{red}{0.0012} & 0.0014 & \textcolor{blue}{0.0014} & 0.0022 \\
    DBL~\cite{gharbi2017deep}/expertA                                          & 0.0017 & \textcolor{blue}{0.0011} & \textcolor{red}{0.0010} & 0.0015 & 0.0014 & \textcolor{blue}{0.0007} & \textcolor{red}{0.0006} & 0.0015 \\
    DBL~\cite{gharbi2017deep}/expertB                                          & 0.0016 & \textcolor{blue}{0.0010} & \textcolor{red}{0.0008} & 0.0014 & 0.0011 & \textcolor{blue}{0.0006} & \textcolor{red}{0.0004} & 0.0014 \\
    IID~\cite{bell2014intrinsic}/reflectance                                   & 0.0015 & \textcolor{blue}{0.0008} & \textcolor{red}{0.0007} & 0.0012 & 0.0011 & \textcolor{red}{0.0006} & \textcolor{blue}{0.0006} & 0.0006 \\
    IID~\cite{bell2014intrinsic}/shading                                       & 0.0014 & \textcolor{blue}{0.0008} & \textcolor{red}{0.0008} & 0.0013 & 0.0008 & \textcolor{blue}{0.0004} & \textcolor{red}{0.0004} & 0.0017 \\
    CycleGAN~\cite{zhu2017unpaired}/photo2ukiyoe                               & 0.0024 & \textcolor{blue}{0.0019} & \textcolor{red}{0.0015} & 0.0013 & 0.0017 & \textcolor{blue}{0.0013} & \textcolor{red}{0.0010} & 0.0018 \\
    CycleGAN~\cite{zhu2017unpaired}/photo2vangogh                              & 0.0026 & \textcolor{blue}{0.0026} & \textcolor{red}{0.0019} & 0.0014 & 0.0020 & \textcolor{blue}{0.0020} & \textcolor{red}{0.0015} & 0.0020 \\
    Colorization~\cite{zhang2016imagecolorization}                             & 0.0016 & \textcolor{blue}{0.0011} & \textcolor{red}{0.0008} & 0.0013 & 0.0010 & \textcolor{blue}{0.0006} & \textcolor{red}{0.0004} & 0.0012 \\
    Colorization~\cite{waechter2014let}                                        & 0.0015 & \textcolor{blue}{0.0009} & \textcolor{red}{0.0007} & 0.0013 & 0.0010 & \textcolor{blue}{0.0005} & \textcolor{red}{0.0003} & 0.0012 \\
    \hline
    Average                               & 0.0027 & \textcolor{blue}{0.0027} & \textcolor{red}{0.0021} & 0.0018* & \textcolor{blue}{0.0018} & 0.0019 & \textcolor{red}{0.0015} & 0.0018* \\
    \hline
    \end{tabular}}
    \caption{\textbf{Quantitative comparison of Temporal Warping Error.} The comparison presented in this table is limited to the results produced by~\cite{bonneel2015blind, lai2018learning} and the proposed method. Lower warp error suggests better temporal consistency. "*" Indicates partially averaged results.}
\vspace{-10pt}
\label{quant_error_table}
\end{table}%

The quantitative results are evaluated on the basis of temporal warp error. \tabref{quant_error_table} shows the quantitative result produced by the methods proposed in \cite{bonneel2015blind, lai2018learning, lei2020blind} and our method. Please note that there are over 830 video sequences in the evaluation datasets and over 115,000 frames. 
Due to the extremely tedious and resource-intensive nature of DVP~\cite{lei2020blind}, only a portion of the evaluation datasets could be processed and the evaluated metrics in the DVP~\cite{lei2020blind} column are averaged over only the processed videos for each task. In order to have a fair comparison, the metrics evaluated on the videos processed with DVP~\cite{lei2020blind} are excluded from the large-scale quantitative comparison and are only presented for reference only. Furthermore, it is also noteworthy that in their paper~\cite{lei2020blind}, the authors do not evaluate the results on the entirety of the dataset and sample tasks for comparative results and report that their warp error is comparable to that of Bonneel et al.~\cite{bonneel2015blind}. 

The lower temporal warp error represents better temporal consistency. The lowest temporal warp error is highlighted in red, and the second best is highlighted in blue. It is worth mentioning that the temporal warp error does not take into account the perceptual quality of the produced frames and assigns a lower value to blurry videos (as produced by~\cite{lei2020blind}). The quality of the results produced by each of the discussed methods can be verified through the accompanying supplementary video. 
It is evident from \tabref{quant_error_table} that the proposed model produces state-of-the-art consistency results. Please note that, unlike the proposed method, all of the compared methods require the availability of raw/unprocessed videos for restoring temporal dynamics at both the training and test time. 

We further verify the efficacy of the proposed model through a specialized metric (as presented in~\cite{jin2019msm}) for evaluating motion smoothness in videos, on the portion of evaluation datasets processed through the DVP~\cite{lei2020blind} and present our findings in the accompanying supplemental.

It is worth mentioning that some attempts to judge the  perceptual quality through metrics like mean PSNR (performance degradation~\cite{lei2020blind}) have been proposed for this task. We intentionally disregarded this metric and argue that this metric might be helpful to evaluate the performance of successive iterations of their prior-based method where each iteration leads to progressively better results, but it can be a bit misleading in judging the overall perceptual quality of restored videos for this task. A similar phenomenon can be observed with other local perceptual metrics such as LPIPS~\cite{zhang2018unreasonable} as it is impossible to quantize a meaningful deviation of perceptual distance or PSNR from per-frame processed videos due to the lack of ground truth data.


In order to be consistent with the prior works, we nonetheless present the LPIPS~\cite{zhang2018unreasonable} results in~\ref{quant_lpips_table}. The proposed method performs competitively with the method proposed by~\cite{lai2018learning} while achieving better temporal consistency. It is noteworthy that each of the discussed metrics does not take into account the aspects considered by the other metric and only provides limited insights about the results. Therefore, to properly evaluate the models on both perceptual quality and temporal consistency, thorough user studies are essential and are used as the primary metric for the evaluation of this task. We conducted two separate user studies to properly evaluate all of the discussed methods and present our findings in the next sub-section.

\begin{table}[h]
\adjustbox{max width=0.48\textwidth}{
  \centering
    \begin{tabular}{|l|ccc|c|ccc|c|}
    \hline
    \multicolumn{1}{|c|}{\multirow{2}[2]{*}{Task}} & \multicolumn{4}{c|}{DAVIS}    & \multicolumn{4}{c|}{VIDEVO} \\
    \multicolumn{1}{|c|}{} & Bonneel\cite{bonneel2015blind} & Lai\cite{lai2018learning}   & \multicolumn{1}{c}{Ours} & DVP\cite{lei2020blind}* & Bonneel\cite{bonneel2015blind} & Lai\cite{lai2018learning}   & \multicolumn{1}{c}{Ours} & DVP\cite{lei2020blind}* \\
    \hline
    WCT~\cite{li2017universal}/antimono                                        & 0.2166 & \textcolor{red}{0.0431} & \textcolor{blue}{0.0485} & 0.3394 & 0.2180 & \textcolor{blue}{0.0451} & \textcolor{red}{0.0447} & 0.2992 \\
    WCT~\cite{li2017universal}/asheville                                       & 0.2056 & \textcolor{red}{0.0373} & \textcolor{blue}{0.0449} &   --    & 0.2680 & \textcolor{blue}{0.0514} & \textcolor{red}{0.0485} & -- \\
    WCT~\cite{li2017universal}/candy                                           & 0.2588 & \textcolor{red}{0.0435} & \textcolor{blue}{0.0497} &   --    & 0.2175 & \textcolor{blue}{0.0534} & \textcolor{red}{0.0514} & -- \\
    WCT~\cite{li2017universal}/feathers                                        & 0.2096 & \textcolor{red}{0.0467} & \textcolor{blue}{0.0508} &   --    & 0.1519 & \textcolor{blue}{0.0590} & \textcolor{red}{0.0540} & -- \\
    WCT~\cite{li2017universal}/sketch                                          & 0.1451 & \textcolor{red}{0.0499} & \textcolor{blue}{0.0511} &   --    & 0.1871 & \textcolor{blue}{0.0517} & \textcolor{red}{0.0496} & -- \\
    WCT~\cite{li2017universal}/wave                                            & 0.1777 & \textcolor{red}{0.0444} & \textcolor{blue}{0.0485} & 0.3423 & 0.2583 & \textcolor{red}{0.0749} & \textcolor{blue}{0.0793} & 0.3451 \\
    Fast-neural-style~\cite{johnson2016perceptualloss}/princess                & 0.2238 & \textcolor{red}{0.0646} & \textcolor{blue}{0.0744} & 0.4782 & 0.1610 & \textcolor{red}{0.0555} & \textcolor{blue}{0.0625} & 0.2566 \\
    Fast-neural-style~\cite{johnson2016perceptualloss}/udnie                   & 0.1473 & \textcolor{red}{0.0503} & \textcolor{blue}{0.0609} & 0.3660 & 0.0800 & \textcolor{red}{0.0514} & \textcolor{blue}{0.0604} & 0.1487 \\
    DBL~\cite{gharbi2017deep}/expertA                                          & 0.0795 & \textcolor{red}{0.0379} & \textcolor{blue}{0.0495} & 0.1453 & 0.0742 & \textcolor{red}{0.0504} & \textcolor{blue}{0.0584} & 0.1349 \\
    DBL~\cite{gharbi2017deep}/expertB                                          & 0.0530 & \textcolor{red}{0.0359} & \textcolor{blue}{0.0481} & 0.1361 & 0.1359 & \textcolor{blue}{0.0639} & \textcolor{red}{0.0578} & 0.1947 \\
    IID~\cite{bell2014intrinsic}/reflectance                                   & 0.1011 & \textcolor{red}{0.0493} & \textcolor{blue}{0.0536} & 0.1389 & 0.0783 & \textcolor{red}{0.0659} & \textcolor{blue}{0.0731} & 0.2635 \\
    IID~\cite{bell2014intrinsic}/shading                                       & 0.0646 & \textcolor{red}{0.0527} & \textcolor{blue}{0.0643} & 0.1558 & 0.1191 & \textcolor{red}{0.0506} & \textcolor{blue}{0.0592} & 0.2302 \\
    CycleGAN~\cite{zhu2017unpaired}/photo2ukiyoe                               & 0.0908 & \textcolor{red}{0.0394} & \textcolor{blue}{0.0502} & 0.2351 & 0.1462 & \textcolor{red}{0.0505} & \textcolor{blue}{0.0575} & 0.2571 \\
    CycleGAN~\cite{zhu2017unpaired}/photo2vangogh                              & 0.1230 & \textcolor{red}{0.0390} & \textcolor{blue}{0.0494} & 0.2565 & 0.1224 & \textcolor{red}{0.0519} & \textcolor{blue}{0.0551} & 0.1701 \\
    Colorization~\cite{zhang2016imagecolorization}                             & 0.0616 & \textcolor{red}{0.0402} & \textcolor{blue}{0.0468} & 0.1555 & 0.0867 & \textcolor{red}{0.0495} & \textcolor{blue}{0.0566} & 0.1441 \\
    Colorization~\cite{waechter2014let}                                        & \textcolor{red}{0.0329} & \textcolor{blue}{0.0373} & 0.0475 & 0.1474 & 0.1317 & \textcolor{red}{0.0081} & \textcolor{blue}{0.0220} & 0.1558 \\
    \hline
    Average                                           & 0.1369 & \textcolor{red}{0.0445} & \textcolor{blue}{0.0524} & 0.2414* & 0.1523 & \textcolor{red}{0.0521} & \textcolor{blue}{0.0556} & 0.2167* \\
    \hline
    \end{tabular}}%
    \caption{\textbf{Quantitative comparison of LPIPS~\cite{zhang2018unreasonable}.} The comparison presented in this table is limited to the results produced by~\cite{bonneel2015blind, lai2018learning} and the proposed method. Lower LPIPS suggests better perceptual similarity. "*" Indicates partially averaged results.}
\vspace{-22pt}
\label{quant_lpips_table}
\end{table}%

\subsubsection{User Study}


\begin{figure}[h]
\begin{center}
\includegraphics[width=0.48\linewidth]{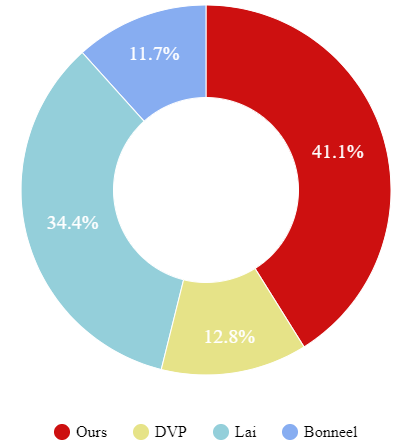}
\end{center}
\vspace{-0.05\linewidth}
\caption{\textbf{User Study.} This donut chart highlights the user preferences collected through the user study. A majority of the participants preferred the results produced by the proposed method.} 
\vspace{-8pt}
\label{fig_user_study}
\end{figure}
We conducted $2$ separate user studies; a comprehensive and a factorized user study, to properly evaluate the performance of the proposed model. The first user study consisted of $36$ participants with $150$ different scenes processed through each of the proposed models. Each participant was asked to judge $5$ videos processed through each of the proposed methods for this task. The users were shown a randomly sampled scene from the video pool processed through~\cite{bonneel2015blind, lai2018learning, lei2020blind} and this work, simultaneously. The participants were instructed to judge the videos on the basis of naturalness, quality, consistency, content, and style preservation. Figure \ref{fig_user_study} presents the findings of the first (comprehensive) user study. On average, 41\% of the users preferred the videos restored by the proposed method. Furthermore, the results generated by the method proposed in Lai et al.~\cite{lai2018learning} also received a significant preference in the user study but it is worth noting that all the previously proposed approaches requires the unprocessed counterparts of the frame-wise processed videos at test time to restore the temporal consistency, whereas the proposed model restores the temporal consistency by only taking into account the inconsistent video at the test time. More details about the environment of the user studies and the factorized study are presented in the accompanied supplemental.



\subsection{Qualitative Results}
\vspace{-15pt}
\begin{figure}[h]
\begin{center}
\includegraphics[width=1.0\linewidth]{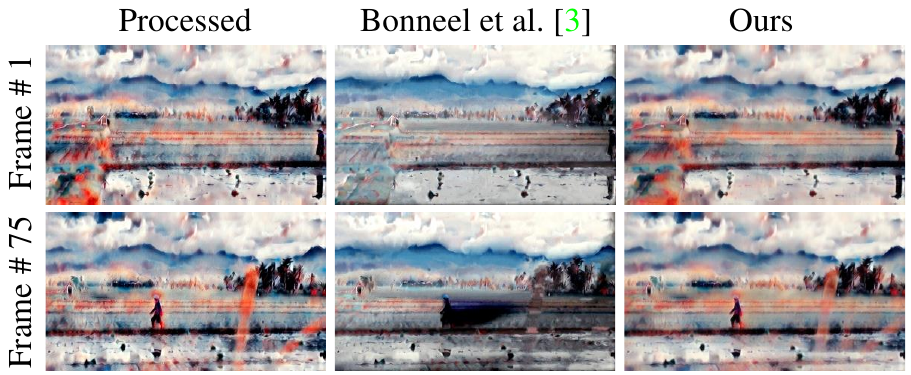}
\end{center}
\vspace{-15pt}
\caption{\textbf{Comparison with Bonneel et al.~\cite{bonneel2015blind}} The method proposed by Bonneel et al.~\cite{bonneel2015blind} has difficulty in handling occlusion and long term consistency and produce temporal artifacts.} 
\vspace{-15pt}
\label{bonvsours}
\end{figure}
\begin{figure}[ht]
\begin{center}
\includegraphics[width=1.0\linewidth]{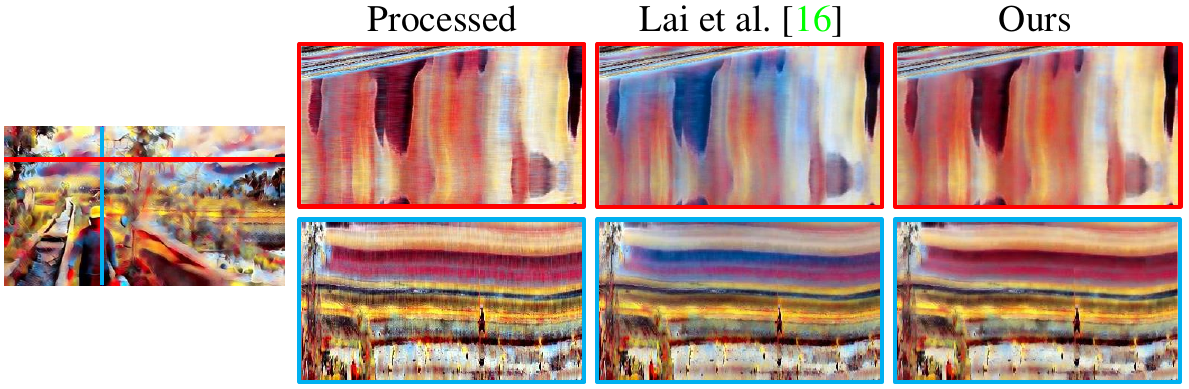}
\end{center}
\vspace{-15pt}
\caption{\textbf{Comparison with Lai et al.~\cite{lai2018learning}} This figure produces epipolar temporal comparison of videos restored by Lai et al.~\cite{lai2018learning} and the proposed method. The method proposed in~\cite{lai2018learning}, due to its dependence on raw videos and implicit siphoning/mixing of motion dynamics from the raw video introduces bias towards the raw videos.} 
\vspace{-15pt}
\label{laivsour}
\end{figure}

Figures~\ref{bonvsours} $\sim$~\ref{laivsour} present some of the comparative qualitative results produced by the methods proposed in~\cite{bonneel2015blind, lai2018learning, lei2020blind} and the proposed model. The results produced by~\cite{bonneel2015blind} fail to retain the perceptual quality of video sequences where occlusion and deocclusion occur. Whereas the frames processed by DVP~\cite{lei2020blind} are significantly blurry and lose the translation effect. In contrast to the comparison with DVP~\cite{lei2020blind} and Boneel et al.~\cite{bonneel2015blind}, the comparison with the results produced through the method proposed by Lai et al.~\cite{lai2018learning} can be a bit tricky. In order to aid the readers, we provide an epipolar comparison between the frames restored through their method and ours in Fig.~\ref{laivsour}. Due to the implicit definition of motion dynamics in their method~\cite{lai2018learning}, their method fails to faithfully retain the perceptual quality in the restored frames and often flips the colors/appearance of objects in the restored video according to the raw video. This bias towards the raw videos is quite prominent in methods like DVP~\cite{lei2020blind} and Lai et al.~\cite{lai2018learning}. Contrasting to their formulation, the explicit design of motion dynamics in the proposed method mitigates the need for raw videos at test time which consequently eliminates this bias and enables the proposed method to faithfully restore videos. Further qualitative comparisons and higher quality results, along with the results on super resolution~\cite{ledig2017sisr}, inpainting~\cite{eunhye2021inpainting}, multiple restoration iterations and further user studies are presented in the supplemental.

\begin{figure}[ht]
\begin{center}
\includegraphics[width=1.0\linewidth]{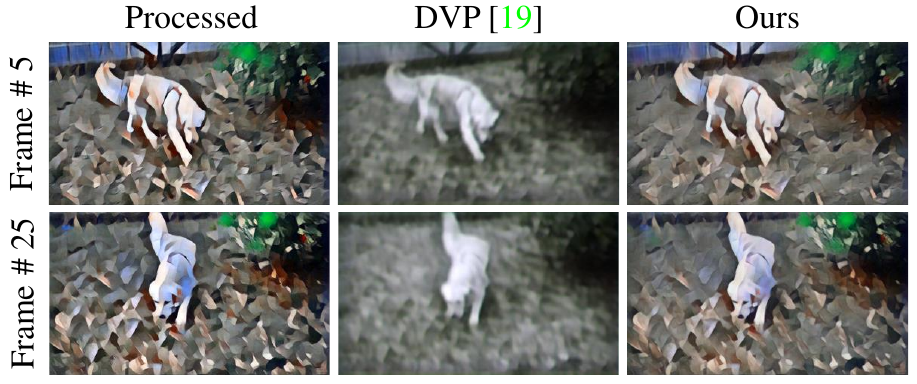}
\end{center}
\vspace{-10pt}
\caption{\textbf{Comparison with DVP~\cite{lei2020blind}} DVP~\cite{lei2020blind} produces blurry frames and compromises the translation effect. This phenomenon is prominent in videos where new content is generated e.g. style transfer~\cite{johnson2016perceptualloss}, CycleGAN~\cite{zhu2017unpaired}.} 
\vspace{-10pt}
\label{dvpvsour}
\end{figure}

\section{Limitations and Future work}
\vspace{-5pt}
The proposed model has difficulties in restoring videos with very low frame rate and in the case of style transferred videos, some finer inconsistencies like finer brush strokes are lost, which makes the resulting video seem a little dull (which is to be expected from all the proposed approaches for this task). A good direction for future work in this field can be the direct denoising of conventional optical flow with deterministic warp operators. The developed consistent flow models can also be used for a number of video processing applications like denoising and dehazing.
\vspace{-5pt}
\section{Conclusion}
\vspace{-2pt}
In this work, we present a task-agnostic temporal consistency correction framework that restores natural video dynamics by inferring and utilizing consistent motion representations from temporally inconsistent videos. The proposed strategy 
mitigates the need for siphoning video dynamics from the unprocessed videos at test time consequently mitigating the compromise in translation effects that is prominant in the currently proposed approaches for this task and expands the scope of its employ-ability to a wider set of applications. 
Through extensive experimentation and user studies, we demonstrate that despite the limited access to data at test-time the proposed method compares favorably to the pre-existing methods available for this task and achieves SOTA performance.

\section*{Acknowledgements}
This work was supported by Institute of Information \& communications Technology Planning \& Evaluation (IITP) grant funded by the Korea government(MSIT) (No.2022-0-00156, Fundamental research on continual meta-learning for quality enhancement of casual videos and their 3D metaverse transformation) and Samsung Electronics Co., Ltd, and Samsung Research Funding Center of Samsung Electronics under Project Number SRFCIT1901-06.


{\small
\bibliographystyle{ieee_fullname}
\bibliography{main}
}

\end{document}